\pdfoutput=1

\documentclass[11pt]{article}

\usepackage{coling}
\usepackage{multirow}
\usepackage{times}
\usepackage{latexsym}
\usepackage[T1]{fontenc}

\usepackage[utf8]{inputenc}
\usepackage{listings,float}
\usepackage{microtype}
\lstdefinestyle{promptstyle}{
    backgroundcolor=\color{lightgray!50},
    basicstyle=\ttfamily\small\color{black},
    breaklines=true,
    breakindent=0pt,
    frame=single,
    framesep=5pt,
    numbers=none,
    prebreak={},
    postbreak={},
    xleftmargin=5pt,
    xrightmargin=5pt,
    belowskip=10pt,
    aboveskip=10pt,
    columns=fullflexible,
    keepspaces=true
}
\usepackage{inconsolata}
\usepackage{graphicx}
\usepackage{amssymb}
\usepackage{bbding}
\usepackage{amsmath}
\usepackage{array}

\usepackage{graphicx}
\usepackage{algorithm}
\usepackage{algorithmic}
\usepackage{booktabs}
\usepackage{multirow}
\usepackage{amsfonts}
\usepackage{amssymb}

\usepackage{nicefrac}
\usepackage{url}
\usepackage{xcolor}
\usepackage{colortbl}
\usepackage{makecell}

\makeatletter
\newcommand{\ssymbol}[1]{$^{\@fnsymbol{#1}}$}
\makeatother

\newcommand{\liu}[1]{{\color{black}#1}}

\title{SLaVA-CXR: Small Language and Vision Assistant for \\ Chest X-ray Report Automation}

\author{Jinge Wu\textsuperscript{1}\thanks{These authors contributed equally to this work.}, 
        Yunsoo Kim\textsuperscript{1}\footnotemark[1], 
        \textbf{Daqian Shi\textsuperscript{1}}, 
        \textbf{David Cliffton\textsuperscript{2}}, \\ 
        \textbf{Fenglin Liu\textsuperscript{2}\thanks{Corresponding authors.}},
        \textbf{Honghan Wu}\textsuperscript{1}\textsuperscript{3}\footnotemark[2]\\
\textsuperscript{1}University College London 
\textsuperscript{2}University of Oxford\\
\textsuperscript{3}University of Glasgow\\
\texttt{\{jinge.wu.20, honghan.wu\}@ucl.ac.uk}\\ 
\texttt{\{fenglin.liu\}@eng.ox.ac.uk}\\ 
}

\begin{document}
\maketitle
\begin{abstract}
Inspired by the success of large language models (LLMs), there is growing research interest in developing LLMs in the medical domain to assist clinicians. However, for hospitals, using closed-source commercial LLMs involves privacy issues, and developing open-source public LLMs requires large-scale computational resources, which are usually limited, especially in resource-efficient regions and low-income countries. We propose an open-source Small Language and Vision Assistant (SLaVA-CXR) that can be used for Chest X-Ray report automation.
To efficiently train a small assistant, we first propose the Re$^3$Training method, which simulates the cognitive development of radiologists and optimizes the model in the `\textit{Re}cognition', `\textit{Re}asoning', and `\textit{Re}porting' training manner.
Then, we introduce a data synthesis method, RADEX, which can generate a high-quality and diverse training corpus with privacy regulation compliance.
The extensive experiments show that our SLaVA-CXR built on a 2.7B backbone not only outperforms but also achieves 6 times faster inference efficiency than previous state-of-the-art larger models
\footnote{https://github.com/knowlab/SLaVA-CXR}.
\end{abstract}

\section{Introduction}
\label{sec:intro}

In recent years, the integration of artificial intelligence into medical imaging has advanced diagnostics and patient care, particularly as assistance tools. Notable \liu{large language models (LLMs)} include \liu{GPT-4}, GPT-4-Vision \cite{gpt4}, and LLaVA \cite{liu2023visual}, which demonstrate impressive performance in general domain tasks and show promising performance in medical \liu{(vision and)} question answering \cite{zhou2023survey}, as in \liu{MedPaLM \cite{singhal2022large}}, LLaVA-Med \cite{li2023llava} and Med-PaLM M \cite{tu2023towards}.

However, these LLMs encounter limitations that hinder their practical application in real-world medical data involving patient information. For instance, assistants like ChatGPT utilizing the GPT-4-Vision model or similar proprietary API services raise concerns regarding the privacy of patient information. To avoid any release of patient information, some hospitals adopt a cautious approach by storing the data in an environment with restricted intranet access and no internet connection \cite{basil2022health,basu2020restoring}. Even for hospitals storing data with internet access, the API services must adhere to strict regulations such as the Health Insurance Portability and Accountability Act \cite{hipaa2024}. In other words, hospitals must de-identify clinical notes and establish secure connections to mitigate the risk of privacy breaches, which limits the usage of such services.

While proprietary models may face privacy concerns, open-source \liu{LLMs} such as LLaVA \cite{liu2023visual} provide the advantage of local usage even within an offline environment. However, it is essential to note that even open-source models, despite their greater accessibility, their relatively high computational resource demands, and lower performance in comprehending medical knowledge still remain unresolved \cite{jin2024hidden}. Meeting these demands proves to be a significant hurdle in many hospitals, highlighting the need for work to make an efficient model with improved performance for a medical imaging assistant.
Moreover, when adapting LLMs to the medical domain, most existing works adopt the public MIMIC-III and MIMIC-IV \cite{johnson2016mimic,johnson2023mimic} datasets for training. However, access to these data is restricted to credentialed individuals with CITI training \cite{CITI}. This restriction extends to any models/products derived from these datasets, including synthetic data or generative models trained using them.


To this end, we first propose an efficient training method Re$^3$Training, which employs a strategically layered approach, systematically deepening the model's expertise from fundamental visual comprehension to advanced clinical articulation. Mirroring the cognitive development of radiologists, the training pipeline comprises three sequential stages: Radiological Pattern Recognition Study (Recognition), Diagnostic Reasoning with Instruction Tuning (Reasoning), and Clinical Reporting Learning (Reporting).
(1) The recognition stage lays the foundational groundwork, focusing on basic visual feature extraction and medical concept alignment. (2) Building upon this, the Reasoning stage elevates the model's capabilities to interpret these patterns within a diagnostic framework. (3) Finally, the Reporting stage hones the model's ability to synthesize its learned knowledge into coherent, professional-grade radiology reports. This progressive intensification of skills ensures that each subsequent stage leverages and extends the competencies acquired in the previous one, ultimately yielding a model proficient in generating comprehensive and accurate radiology reports.

Then, we further introduce an efficient data synthesis method RADEX, which can synthesize a high-quality and diverse training corpus (RADiology EXpertise corpus) from publicly available clinical-standard case reports and X-ray image pairs. These are sourced from radiopaedia.org, which is publicly accessible and free of privacy concerns.
We use the synthesized RADEX to train SLaVA-CXR, a small but efficient vision and language assistant specifically tailored for CXR report automation. Our extensive experiments show that our SLaVA-CXR achieves the best performance in CXR report generation and summarization compared to existing state-of-the-art LLMs.

Our paper makes the following contributions:
\begin{itemize} 
    \item  We propose a small vision and language assistant SLaVA-CXR, which outperforms larger models with a maximum of 6 times faster inference time.
 
    \item We propose an efficient training method Re$^3$Training and an efficient data synthesis method RADEX to enable our SLaVA-CXR to accurately comprehend complex medical data, achieving superior performance.
    
    \item We perform extensive experiments on two standard benchmarks and further invite medical experts to conduct human evaluation to prove the effectiveness of our model.

\end{itemize}

\begin{figure*}[]
\begin{center}
\centerline{\includegraphics[width=1.9\columnwidth,trim={32 90 50 80},clip]{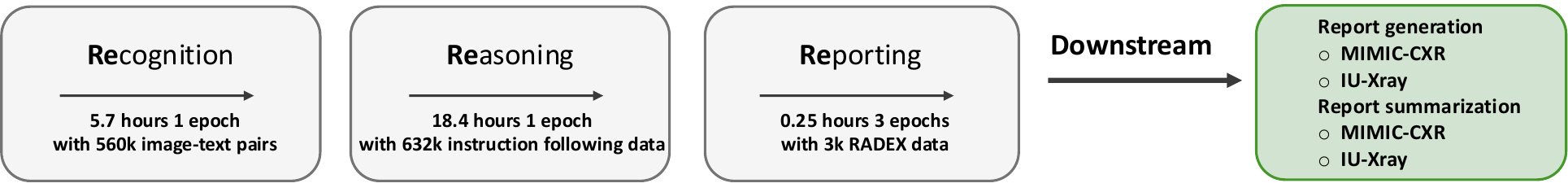}}
\vspace{-7pt}
\caption{The proposed Re$^3$Training pipeline. Stage 1: The Recognition stage aims to improve model capacity in aligning clinical concepts between two modalities; Stage 2: The Reasoning stage aims to capture CXR image nuances; Stage 3: The Reporting stage is for learning the clinical notes of CXRs. }
\label{fig:training}
\end{center}
\vspace{-20pt}
\end{figure*}
\begin{table*}[t]
\scriptsize
\centering
\begin{tabular}{lllccc c}
\toprule
\textbf{Dataset} & \textbf{Size} &  \textbf{Description} & \textbf{Stage 1} & \textbf{Stage 2} & \textbf{Stage 3}   & \textbf{Evaluaion}\\ \hline
Blip\_Laion\_CC\_SBU  & 558,128 &   Images and instructions         &   \Checkmark     &    -     &      -  &  -   \\ 
CXR-Alignment  & 1,436 &  Images and instructions          &   \Checkmark     &    -     &      -  &  -   \\ 
LLaVA-665k-Subset  & 624,603 &   Images and instructions         &    -    &    \Checkmark     &      -  &  -   \\ 
CXR-Instruction  & 7,335 &   Images and instructions         &    -    &    \Checkmark     &      -  &  -   \\ 
CXR Clinical Note  & 3,333 &   Images, reports, and instructions         &   -     &    -     &      \Checkmark  &  -   \\ 
MIMIC-CXR  & 1,732 &   Images and reports         &   -     &    -     &      -  &  \Checkmark  \\
IU-Xray  & 3,301 &   Images and reports         &   -     &    -     &     -  &  \Checkmark  \\ \bottomrule
\end{tabular}
\vspace{-5pt}
\caption{Data collection for each stage. Stage1, Stage 2, and Stage 3 refer to the training stages shown in Figure \ref{fig:training}.}
\label{data_train}
\vspace{-5pt}
\end{table*}

\section{Related Works}
In the critical domain of clinical applications, where precise interpretation of medical images is paramount for accurate diagnosis and patient care, the demand for sophisticated multimodal large language models has become increasingly apparent \nocite{liu2023medical}. Notable contributions in this field include PMC-VQA \cite{zhang2023pmc} and LLaVA-Med \cite{li2023llava}, which have demonstrated promising capabilities in bridging the gap between visual input and textual understanding in medical contexts. However, these models often encounter significant limitations when tasked with generative assignments, such as comprehensive report generation. This shortcoming can be largely attributed to their training paradigm, which predominantly relies on visual question answering datasets extracted from PubMed literature articles, specifically focusing on figure captions and associated legends.

Recent advancements in the field have seen a shift towards addressing the complex task of medical report generation \cite{liu2021PPKED}. Notable examples include CXR-LLaVA \cite{lee2023cxr} and MAIRA-2 \cite{bannur2024maira}, which have made substantial strides in generating detailed and clinically relevant reports from medical images. However, these models are not without their drawbacks. They often rely on intensive computational resources, which can pose significant challenges for real-time clinical applications due to their resource requirements and inference latency. Furthermore, many of these models are trained on proprietary or credentialed datasets, raising concerns about their accessibility and the potential for widespread adoption and validation by the broader community.

\section{\liu{Re$^3$Training}}
Our innovative Re$^3$Training pipeline is designed to systematically develop a model's capabilities in chest X-ray (CXR) report automation. This approach mirrors the cognitive development of radiologists, progressing through three critical stages: Recognition, Reasoning, and Reporting. Each stage builds upon the previous, gradually deepening the model's expertise from fundamental visual comprehension to advanced clinical articulation.

\subsection{\liu{Recognition}}

In the recognition stage, we train the model to associate visual features in chest X-ray images with corresponding medical concepts. This process involves generating captions for image-text pairs, which helps the model learn to describe radiological patterns accurately.
More specifically, our approach focuses on training the projector $P$, which connects the vision encoder $E$ and the Phi-2 language model $L$. We keep $E$ and $L$ frozen during this phase to preserve their pre-trained knowledge while allowing $P$ to learn the specific visual-textual mapping required for chest X-ray interpretation. This method enables the model to develop radiological pattern recognition skills without altering its fundamental visual and linguistic capabilities.
We formulate the training objective for generating accurate captions as follows:
\begin{equation}
\footnotesize
\min_{\theta} \mathcal{L}R^{(1)} = -\sum \log p(y|E(x), P{\theta}(E(x)))
\end{equation}
where $\min_{\theta} \mathcal{L}R^{(1)}$ denotes the optimization objective for the first stage (Recognition) of our Re$^3$Training pipeline, aiming to minimize the loss function $\mathcal{L}R^{(1)}$ with respect to the parameters $\theta$ of the projector $P$. $x$ is the input image, $y$ is the corresponding text, and $p$ is the probability distribution over the vocabulary.
To ensure comprehensive learning, we utilize a diverse dataset $\mathcal{D}_{R^{(2)}}$ comprising 560 image-text pairs from  \citet{liu2023visual}, and extract 1,436 CXR-related pairs from \citet{li2023llava}. This curated dataset ensures broad coverage of medical imaging concepts while maintaining a focus on CXR-specific features.

\subsection{\liu{Reasoning}}

The reasoning stage builds upon the pattern recognition skills developed in the previous phase, extending the model's capabilities to include diagnostic reasoning. In this stage, we train the model to interpret chest X-ray images beyond simple feature identification, enabling it to draw clinical implications and suggest potential diagnoses.

In this stage, we fine-tune all components - $E$, $P$, and $L$ - to capture the nuances of CXR images and associated diagnoses. The loss function is:
\begin{equation}
\footnotesize
\mathcal{L}R^{(2)} = \mathcal{L}\text{ce}(y_\text{pred}, y_\text{true}) + \lambda\mathcal{L}_\text{reg}(\theta_E, \theta_P, \theta_L)
\end{equation}
where $\mathcal{L}\text{ce}$ is the cross-entropy loss between predicted and true outputs, $\mathcal{L}\text{reg}$ is a regularization term, and $\lambda$ is a hyperparameter balancing the two terms. The input-output relationship is defined as $y_\text{pred} = L(P(E(x)), i)$, where $x$ is the input image and $i$ is the instruction. 

For this phase, we curate a collection of instruction tuning dataset $\mathcal{D}_{R^{(2)}}$, consisting of 632k samples from \citet{liu2023visual} and CXR-specific instructions extracted from \citet{li2023llava}. This dataset is designed to expose the model to a wide range of diagnostic scenarios and instruction formats, enhancing its ability to reason about CXR findings in various clinical contexts.

\subsection{\liu{Reporting}}

The reporting stage focuses on optimizing the model for CXR report generation and enhancing its ability to follow specific writing instructions. In this final phase, we fine-tune the model to produce radiology reports that are coherent, accurate, and clinically relevant. Our approach addresses two key aspects: the generation of comprehensive CXR reports and the adherence to varied writing instructions. This stage builds upon the diagnostic reasoning skills developed in earlier phases, extending them to include the structured articulation of findings in a format consistent with radiological reporting standards. By integrating these elements, we aim to create a system capable of generating professional-quality reports while adapting to different reporting styles and requirements.
To achieve these objectives, we continue to fine-tune all components with a multi-task objective:
\begin{equation}
\footnotesize
\mathcal{L}R^{(3)} = \alpha_1\mathcal{L}\text{rep} + \alpha_2\mathcal{L}\text{instr} + \alpha_3\mathcal{L}\text{reg}
\end{equation}
where $\mathcal{L}\text{rep}$ is the report generation loss, corresponding to our first key aspect of producing comprehensive CXR reports. $\mathcal{L}\text{instr}$ is the instruction-following loss, addressing our second key aspect of adapting to various writing instructions. $\mathcal{L}\text{reg}$ is a regularization term to prevent overfitting, and $\alpha_1$, $\alpha_2$, $\alpha_3$ are weighting coefficients that balance the importance of each component in the overall objective.
For this phase, we employ the RADiology EXpertise Corpus (RADEX), a custom dataset we developed to support our training objectives. RADEX comprises comprehensive radiological reports, expert discussions, and diverse instructions, providing a rich information source for enhancing the model's diagnostic reasoning capabilities. This dataset is designed to expose the model to real-world clinical scenarios, improving its ability to interpret chest X-ray images accurately. Further details about RADEX, including its composition and creation process, will be discussed in the following.

\begin{figure}[t]
\centering
\centerline{\includegraphics[width=1\linewidth,trim={0 50 0 50},clip]{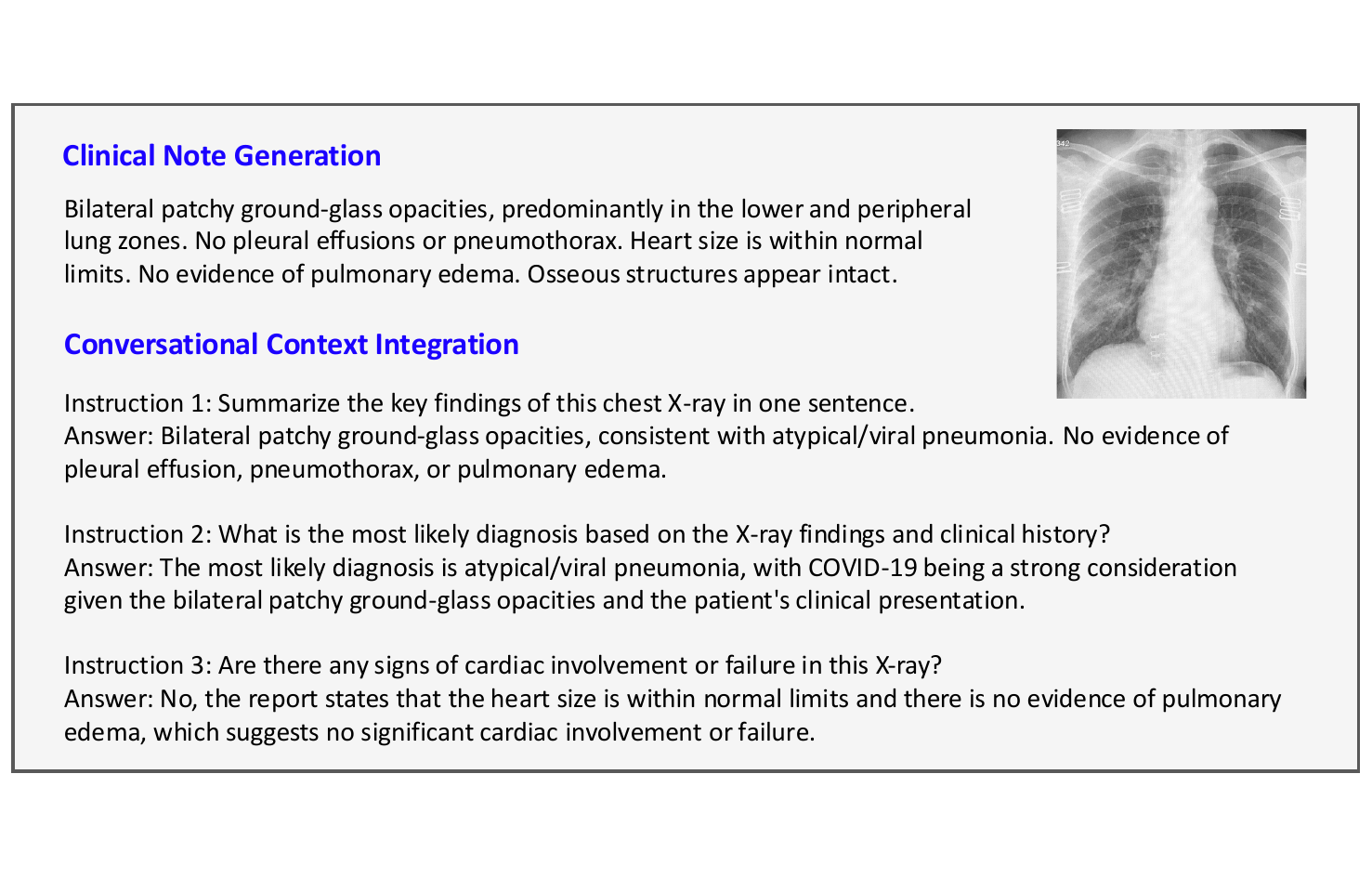}}
\caption{An example of RADEX corpus.}
\label{fig:data_template}
\end{figure}

\section{\liu{RADiology EXpertise Corpus (RADEX)}}


To address limitations in traditional clinical report generation datasets, as shown in Figure~\ref{fig:data_template}, we introduce RADEX, a novel corpus derived from peer-reviewed, open-access radiology case studies. Unlike datasets such as MIMIC-CXR, which often lack comprehensive assessments and may be biased towards severe conditions, RADEX offers a more diverse and holistic representation of radiological cases. This approach aims to enhance model training by providing richer contextual information, including diagnostic rationales and alternative interpretations, thereby improving the quality and breadth of generated reports.


More specifically, our data construction process encompasses the following key components: 
(1) \textit{Synthetic Clinical Note Generation}: Utilizing GPT-4, we generate structured clinical notes that integrate the following components: case description that provides basic background information for the case; case representation of the image that includes the specific imaging findings; case discussion that provides in-depth analysis and diagnostic reasoning for the case. This process ensures a consistent format while preserving the depth and nuance of the original case studies. 
(2) Conversational Context Integration: We enhance the model's instruction-following capabilities by incorporating diverse conversational data. This includes relevant dialogue samples from ~\citet{wu2023towards} and case discussion information from our dataset. By exposing the model to various instruction formats and contextual scenarios, we aim to improve its adaptability in understanding and responding to a wide range of clinical communication styles and requirements.

\begin{table}[t]
\scriptsize
\centering
\begin{tabular}{lll}
\toprule
\textbf{Model}                                    & \textbf{Vision Encoder}          &  \textbf{LLM}    \\ \midrule
LLaVAv0                      &     CLIP ViT-L/14      &  Vicuna-7B  \\
LLaVAv1.5                    &     CLIP ViT-L/14-336px      &    LLaMA2-7B\\
LLaVA-Med                         &   CLIP ViT-L/14        &  Vicuna-7B  \\
TinyGPT-V                            &   EVA-CLIP ViT-g/14        &  Phi-2-2.7B  \\
LLaVA\_phi                      &   CLIP ViT-L/14-336px        &  Phi-2-2.7B \\ \midrule
SLaVA-CXR   &           CLIP ViT-L/14-336px        &  Phi-2-2.7B  \\  \bottomrule
\end{tabular}
\caption{Description of baseline models.}
\label{tab:desc_model}
\end{table}

\begin{table*}[t]
\centering
\scriptsize
\setlength{\tabcolsep}{3pt}

\begin{tabular}{@{}lcccccccccccccccc@{}}
\toprule

  \multirow{2}{*}[-3pt]{\textbf{Methods}}  & \multirow{2}{*}[-3pt]{\textbf{\# Params}} & \multicolumn{7}{c}{\bf MIMIC-CXR} & \multicolumn{7}{c}{\bf IU-Xray}  \\ \cmidrule(lr){3-9}   \cmidrule(lr){10-16}     
  & &    R-L$\uparrow$ &  M$\uparrow$ & B-2$\uparrow$ &   BS$\uparrow$ &  CX$\uparrow$ & RG$\uparrow$ &   RC$\downarrow$ &   R-L$\uparrow$ &  M$\uparrow$ & B-2$\uparrow$ &   BS$\uparrow$ &  CX$\uparrow$ & RG$\uparrow$ &   RC$\downarrow$ \\
\midrule
LLaVAv0 \cite{liu2023visual} & 7B & 7.91  & 15.11 & 3.69 & -15.71 & 9.83  & 2.47 & 2.58          & 0.50           & 12.22          & 1.98          & -19.42         & 14.41          & 2.51          & 2.60 \\

 LLaVA-Med  \cite{li2023llava} & 7B & 8.60  & 16.09 & 3.86 & -15.05 & 9.83  & 4.14 & 2.55          & 2.17           & 13.52          & 2.21          & -18.05         & 14.42          & 5.74          & 2.53 \\
 LLaVAv1.5 \cite{liu2023improved} & 7B & 13.27 & 16.46 & 7.99 & 13.84  & 15.35 & 2.16 & 2.04          & 9.66           & 15.80          & 4.95          & 9.13           & 19.76          & 1.95          & 2.08         \\
  TinyGPT-V \cite{yuan2023tinygpt} & 2.7B & 5.70  & 2.32  & 0.06 & 0.49   & 10.56 & 0.15 & 2.34          & 2.43           & 1.23           & 0.13          & -15.79         & \textbf{25.10} & 0.05          & 2.46         \\
LLaVA\_phi \cite{liu2023visual} & 2.7B & 5.87 & 12.01 & 4.42 & -12.02 & 14.30 & 5.65 & 2.59 & 4.08 & 12.96 & 2.37 & -2.56  & 6.90  & 1.71 & 2.98          \\
 \midrule 
   SLaVA-CXR (Ours)  &  2.7B &  \textbf{13.77} & \textbf{16.79} & \textbf{8.48} & \textbf{23.93}  & \textbf{16.22} & \textbf{8.03} & \textbf{1.79} & \textbf{10.08} & \textbf{17.34} & \textbf{5.81} & \textbf{20.16} & 10.03          & \textbf{5.85} & \textbf{1.94}
\\  
\bottomrule
\end{tabular}
\vspace{-5pt}
    \caption{Chest X-ray report generation performance of methods. R-L, M, B-2, BS, CX, RG, and RC are short for ROUGE-L, METEOR, BLEU-2, BERTScore, CheXbert, RadGraph, and RadCliQ, respectively. Except for RC, all results are reported in percentage (\%).
    $\uparrow$ and $\downarrow$ denote `the higher the better' and `the lower the better', respectively.}
        \label{tab:gene_score}
        \vspace{-5pt}
\end{table*}

\begin{table*}[t]
\centering
\scriptsize
\setlength{\tabcolsep}{3pt}

\begin{tabular}{@{}lcccccccccccccccc@{}}
\toprule

  \multirow{2}{*}[-3pt]{\textbf{Methods}}  & \multirow{2}{*}[-3pt]{\textbf{\# Params}} & \multicolumn{7}{c}{\bf MIMIC-CXR} & \multicolumn{7}{c}{\bf IU-Xray}  \\ \cmidrule(lr){3-9}   \cmidrule(lr){10-16}     
  & &    R-L$\uparrow$ &  M$\uparrow$ & B-2$\uparrow$ &   BS$\uparrow$ &  CX$\uparrow$ & RG$\uparrow$ &   RC$\downarrow$ &   R-L$\uparrow$ &  M$\uparrow$ & B-2$\uparrow$ &   BS$\uparrow$ &  CX$\uparrow$ & RG$\uparrow$ &   RC$\downarrow$ \\
\midrule
LLaVAv0 \cite{liu2023visual} & 7B &  6.90           & 15.83          & 2.32           & -1.56           & 32.74           & 6.14           & 2.07          & 3.78           & 10.40          & 1.12           & -6.67           & 30.90           & 2.63           & 2.22          \\
 LLaVA-Med  \cite{li2023llava} &  7B  &  5.85           & 14.31          & 1.96           & 1.07            & 33.92           & 5.45           & 2.03          & 2.86           & 9.18           & 0.87           & -4.26           & 31.33           & 2.28           & 2.18          \\
 LLaVAv1.5 \cite{liu2023improved} & 7B & 7.87           & 17.38          & 2.54           & 14.26           & 35.16           & 6.51           & 1.78          & 1.21           & 0.72           & 0.22           & -5.71           & \textbf{71.26 } & 0.43           & 1.83          \\
  TinyGPT-V \cite{yuan2023tinygpt} & 2.7B & 4.48           & 1.90           & 0.63           & 5.04            & \textbf{41.96 } & 0.20           & 1.95          & 4.38           & 12.76          & 1.27           & 8.63            & 31.15           & 2.69           & 1.96          \\
LLaVA\_phi \cite{liu2023visual} & 2.7B & 3.63 & 13.21 & 1.17 & 0.02   & 32.08 & 2.42 & 2.88 & 1.05 & 5.22  & 0.21 & -10.05 & 21.58 & 0.05 & 2.51         \\
 \midrule 
   SLaVA-CXR (Ours)  &  2.7B &  \textbf{9.14 } & \textbf{19.92 } & \textbf{3.49 } & \textbf{20.82 } & 35.24           & \textbf{8.47 } & \textbf{1.74} & \textbf{5.08 } & \textbf{14.49 } & \textbf{3.53 } & \textbf{24.17 } & 64.41           & \textbf{3.96 } & \textbf{1.40}
\\  
\bottomrule
\end{tabular}
\vspace{-5pt}
    \caption{Chest X-ray report summarization performance of different methods. }
        \label{tab:sum_score}
        \vspace{-5pt}
\end{table*}

\section{Experiments}
\label{sec:exp}
In this section, we first introduce the settings for our evaluation\footnote{Please refer to our Appendix~\ref{sec:setup} for details.}. We then illustrate the detailed results of our proposed method.

\subsection{Experiment Setup}

\paragraph{Evaluation Data.} The evaluation datasets contain MIMIC-CXR \cite{johnson2019mimic} and IU-Xray \cite{Dina2016IU-Xray} for the task of radiology report generation and summarization. 
\vspace{-5pt}
\noindent\paragraph{Task Description.} We perform the evaluation on CXR report generation and summarization. Each radiology report comprises a ``Findings'' section, detailing the radiologist's observations from the images, and an ``Impressions'' section, summarizing these observations for diagnostic interpretation. In CXR report generation, the objective is to create the "Findings" section from the provided images. For CXR report summarization, the goal is to formulate the ``Impressions'' section using the ``Findings'' and the corresponding images. 
\vspace{-5pt}
\paragraph{Baseline Models.}  
As shown in Table \ref{tab:desc_model}, we choose 5 open-domain models: {LLaVAv0}-7B \cite{liu2023visual}, {LLaVAv1.5}-7B  \cite{liu2023improved}, {LLaVA-Med} \cite{li2023llava}, {TinyGPT-V} \cite{yuan2023tinygpt}, and {LLaVA\_phi} \cite{liu2023visual}. 


\paragraph{Evaluation Metrics.} We evaluate the results with general lexical and radiology-specific metrics. We utilize ROUGE \cite{lin2004rouge}, BLEU \cite{papineni2002bleu}, and METEOR \cite{banerjee2005meteor}, while also incorporating the contextualized evaluation provided by BERTScore \cite{zhang2019bertscore}. In the realm of radiology-specific metrics, we employ tools such as CheXbert \cite{smit2020chexbert}, RadGraph \cite{jain2021radgraph}, and RadCliQ \cite{yu2023evaluating} to gauge the relevance and accuracy of our findings in a medical context.

\begin{table*}[t]
\scriptsize
\centering
\setlength{\tabcolsep}{2pt}
\begin{tabular}{lccccccc}
\toprule
\textbf{Methods}                  & \textbf{No Finding} & \begin{tabular}[c]{@{}c@{}} \bf Enlarged \\ \bf Cardiomediastinum  \end{tabular} & \textbf{Cardiomegaly}  & \textbf{Lung Lesion}   &   \textbf{Lung Opacity}  & \textbf{Edema}   &  \textbf{Consolidation}   \\ \midrule
 LLaVAv0 \cite{liu2023visual} & 51.71          & 50.12          & 49.29          & 50.11          & 50.17          & 51.01          & 50.25 \\ 
 
 LLaVA-Med  \cite{li2023llava} & 51.66          & 50.00          & 53.96          & 54.59          & 50.16          & 52.80          & 52.89   \\ 
 LLaVAv1.5 \cite{liu2023improved} & 49.98          & 50.00          & 50.59          & 50.59          & 50.14          & 50.06          & 50.28 \\
  TinyGPT-V \cite{yuan2023tinygpt} & 51.07          & 49.31          & 49.63          & 51.38          & 53.07          & 50.99          & 52.57  \\
 LLaVA\_phi \cite{liu2023visual}  & 49.10          & 49.97          & 50.67          & 50.04          & 50.50          & 50.37          & 50.44  
\\  \midrule
SLaVA-CXR (Ours) & \textbf{58.87} & \textbf{50.53} & \textbf{57.30} & \textbf{59.15} & \textbf{59.34} & \textbf{59.68} & \textbf{58.15}
\\ \bottomrule
\end{tabular}
\vspace{1pt}
\begin{tabular}{lccccccc}
\toprule
\textbf{Methods}               & \textbf{Pneumonia}    & \textbf{Atelectasis}   & \textbf{Pneumothorax}      & \textbf{Pleural Effusion}    & \textbf{Pleural Other}   & \textbf{Fracture}   & \textbf{Support Devices}   \\ \midrule
         
 LLaVAv0 \cite{liu2023visual} & 51.22          & 50.02          & 51.56          & 50.79          & 50.33          & 49.76          & 53.50 \\ 
 
 LLaVA-Med  \cite{li2023llava}  & 52.86          & 50.00          & 50.00          & 53.47          & 50.16          & 52.52          & 53.73\\

 LLaVAv1.5 \cite{liu2023improved} & 50.41          & 50.00          & 50.00          & 50.43          & 50.11          & 50.10          & \textbf{55.47} \\
  TinyGPT-V \cite{yuan2023tinygpt} & 49.33          & 50.32          & 51.42          & 49.94          & 50.32          & 50.57          & 53.27 \\
 LLaVA\_phi \cite{liu2023visual} & 50.62          & 49.94          & 50.00          & 50.07          & 50.28          & 50.72          & 52.04\\  \midrule
SLaVA-CXR (Ours) & \textbf{55.62} & \textbf{52.21} & \textbf{58.85} & \textbf{55.54} & \textbf{50.54} & \textbf{54.91} & 55.11
\\ \bottomrule
\end{tabular}
\caption{Chest X-ray report classification results of different methods. All results are reported in percentage (\%).}
%
\label{tab:class_auc}
\end{table*}

\subsection{Automatic Evaluation}
\paragraph{Generation Results.}

Table \ref{tab:gene_score} presents a comparison of various models' performance in radiology report generation across multiple datasets. The results reveal that our SLaVA-CXR consistently outperforms other models across a wide range of metrics. 
The results show that larger models such as LLaVA-Med, LLaVAv0, and LLaVAv1.5 did not demonstrate the expected performance gains in these domain-specific tasks, challenging the assumption that increased model size inherently leads to improved performance in specialized domains. Moreover, the medical domain-finetuned variant, LLaVA-Med, fails to distinguish itself significantly in our evaluation metrics.
The discrepancy between LLaVA-Med's performance and that of our SLaVA-CXR model underscores the importance of not only our constructed high-quality data, but also the introduced Re$^3$Training methodology employed in adapting LLMs to specialized medical applications. 

\paragraph{Summarization Results.}
We further report the summarization performance in Table~\ref{tab:sum_score}. As we can see, with fewer parameters, our SLaVA-CXR outperforms previous strong baselines and archives the best results on most metrics.
This observation highlights the need for more nuanced and targeted training strategies when developing models for complex medical imaging interpretation and reporting tasks. TinyGPT-V, while generally lagging behind SLaVA-CXR, shows promise in the CheXbert metric across both tasks, suggesting a potential strength in capturing clinically relevant information despite its compact architecture. These findings underscore the importance of tailored training approaches and architectural considerations in developing effective models for specialized medical tasks, particularly in the realm of radiology report automation.

\paragraph{Classification Results.} To further validate our model's capacity to effectively utilize visual information, we perform the classification task and report the results in Table~\ref{tab:class_auc}.
For a fair comparison, we employ the CheXpert labeler\footnotetext{https://github.com/stanfordmlgroup/chexpert-labeler} to analyze the generated report and compared the classification performance against the original MIMIC-CXR labels. This approach allows us to assess the model's ability to accurately identify and describe clinically relevant findings from CXR images. We calculate the Area Under the Curve (AUC) scores for 14 distinct radiological findings, comparing our SLaVA-CXR model against several baseline models including LLaVA0, LLaVA1.5, LLaVA-Med, TinyGPT-V, and LLaVA\_phi.

The results show that SLaVA-CXR consistently outperforms other models across the majority of findings, showcasing superior classification capabilities. Notably, in detecting critical conditions such as `No Finding', `Edema', and `Lung Opacity', SLaVA-CXR exhibits markedly improved performance compared to its counterparts. This enhanced accuracy is also evident in complex cases that require nuanced interpretation of radiographic features, such as `Enlarged Cardiomediastinum', and `Lung Lesion'. Furthermore, the model's high performance in identifying `Pneumonia' and `Pleural Effusion' underscores its advanced capability in recognizing both parenchymal and pleural abnormalities. The model's success in these areas suggests a sophisticated understanding of the radiographic manifestations of diseases affecting different anatomical compartments of the thorax, from the lung parenchyma to the pleural space.

\begin{figure}[t]
  \centering

    \includegraphics[width=1\linewidth,trim={5 5 5 5},clip]{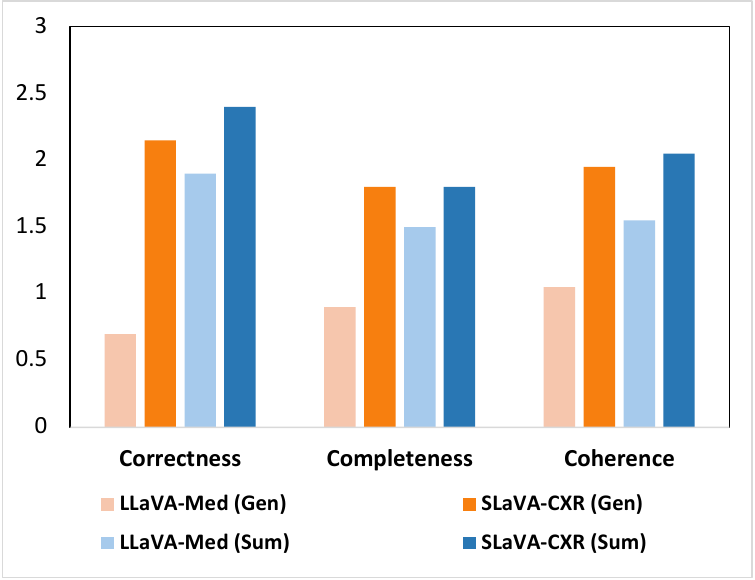} 
   \caption{Human evaluation of our method and LLaVA-Med on the correctness, completeness, and coherence.}
\label{fig:human_eval}
\end{figure}

\begin{table*}[t]
\centering
\scriptsize
\setlength{\tabcolsep}{2pt}

\begin{tabular}{@{}lccccccccccccccccccc@{}}
\toprule

  \multirow{2}{*}[-3pt]{\textbf{Settings}}  &  \multicolumn{3}{c}{\textbf{Re$^3$Training}} & \multicolumn{7}{c}{\bf MIMIC-CXR (Generation)} & \multicolumn{7}{c}{\bf IU-Xray (Generation)}  \\ \cmidrule(lr){2-4} \cmidrule(lr){5-11}   \cmidrule(lr){12-18}     
  & Recognition  & Reasoning & Reporting &   R-L$\uparrow$ &  M$\uparrow$ & B-2$\uparrow$ &   BS$\uparrow$ &  CX$\uparrow$ & RG$\uparrow$ &   RC$\downarrow$ &   R-L$\uparrow$ &  M$\uparrow$ & B-2$\uparrow$ &   BS$\uparrow$ &  CX$\uparrow$ & RG$\uparrow$ &   RC$\downarrow$ \\
\midrule
   Baseline  &- & - & - &5.87 & 12.01 & 4.42 & -12.02 & 14.30 & 5.65 & 2.59 & 4.08 & 12.96 & 2.37 & -2.56  & 6.90  & 1.71 & 2.98\\  \midrule
 
 (a) &\Checkmark & - & -  &  8.04  & 14.80  & 4.74 & -12.02 & 15.10 & 6.30 & 2.42& 4.69 & 13.02 & 2.39 & -13.48 & 7.14 & 3.29 & 2.56 \\
 (b) &\Checkmark & \Checkmark & - & 11.74 & 12.76 & 4.74 & -12.02 & 15.10 & 6.30 & 2.42   & 9.46 & 13.42 & 3.25 & 13.43  & 8.96 & 2.80  & 2.10 \\  \midrule
 
 SLaVA-CXR (Ours)  &  \Checkmark & \Checkmark & \Checkmark &  \textbf{13.77} & \textbf{16.79} & \textbf{8.48} & \textbf{23.93}  & \textbf{16.22} & \textbf{8.03} & \textbf{1.79} & \textbf{10.08} & \textbf{17.34} & \textbf{5.81} & \textbf{20.16} & 10.03          & \textbf{5.85} & \textbf{1.94} \\  
 
\bottomrule
\end{tabular}

\vspace{1pt}

\begin{tabular}{@{}lccccccccccccccccccc@{}}
\toprule

  \multirow{2}{*}[-3pt]{\textbf{Settings}}  &  \multicolumn{3}{c}{\textbf{Re$^3$Training}} & \multicolumn{7}{c}{\bf MIMIC-CXR (Summarization)} & \multicolumn{7}{c}{\bf IU-Xray (Summarization)}  \\ \cmidrule(lr){2-4} \cmidrule(lr){5-11}   \cmidrule(lr){12-18}     
  & Recognition  & Reasoning & Reporting &   R-L$\uparrow$ &  M$\uparrow$ & B-2$\uparrow$ &   BS$\uparrow$ &  CX$\uparrow$ & RG$\uparrow$ &   RC$\downarrow$ &   R-L$\uparrow$ &  M$\uparrow$ & B-2$\uparrow$ &   BS$\uparrow$ &  CX$\uparrow$ & RG$\uparrow$ &   RC$\downarrow$ \\
\midrule
  Baseline  &- & - & - & 3.63 & 13.21 & 1.17 & 0.02   & 32.08 & 2.42 & 2.88 & 1.05 & 5.22  & 0.21 & -10.05 & 21.58 & 0.05 & 2.51\\  \midrule
 
 (a) &\Checkmark & - & -  & 3.73          & \textbf{7.89} & 1.18          & -10.68         & 26.51          & \textbf{2.78} & 2.33          & 1.34          & 5.62          & 0.65          & -9.31          & 26.76          & 1.02 & 2.32  \\
 (b) &\Checkmark & \Checkmark & -  & 5.66          & 7.12          & 1.14          & 18.32          & 32.61          & 2.65          & 1.76          & 2.99          & 5.93          & 2.78          & 19.59          & 52.91          & 1.11 & 1.59 \\  \midrule

 SLaVA-CXR (Ours)  &  \Checkmark & \Checkmark & \Checkmark &  \textbf{9.14 } & \textbf{19.92 } & \textbf{3.49 } & \textbf{20.82 } & 35.24           & \textbf{8.47 } & \textbf{1.74} & \textbf{5.08 } & \textbf{14.49 } & \textbf{3.53 } & \textbf{24.17 } & 64.41           & \textbf{3.96 } & \textbf{1.40}\\  
\bottomrule
\end{tabular}
    \caption{Ablation study of our SLaVA-CXR, which introduces a Re$^3$Training pipeline, which include `Recognition', `Reasoning', and `Reporting' training stages. The `Reporting' training stage further introduces the proposed RADEX data synthesis method.}
        \label{tab:ablation_study}
\end{table*}
\begin{table}[t]
\scriptsize
\centering

\setlength{\tabcolsep}{2pt}
\begin{tabular}{lccccc}
\toprule
  \multirow{2}{*}[-3pt]{\textbf{Methods}}        & \multicolumn{2}{c}{\textbf{Generation}} & \multicolumn{2}{c}{\textbf{Summarization}} &    \multirow{2}{*}[-3pt]{\textbf{Average}}    \\ \cmidrule(lr){2-3} \cmidrule(lr){4-5}  
   
   & MIMIC-CXR & IU-Xray  & MIMIC-CXR & IU-Xray \\ \midrule
LLaVA-Med &  17.56 & 19.59 & 9.79 & 11.64 & 15.50 \\
LLaVAv1.5   &  5.38 & 5.62 & 7.20 & 4.97 & 5.79\\ \midrule

SLaVA-CXR &   \textbf{3.32} & \textbf{4.45} & \textbf{1.26} & \textbf{1.10} & \textbf{2.53}\\  \bottomrule
\end{tabular}

\caption{Inference efficiency comparison (seconds per instance) for each task. We compare our approach with two previous state-of-the-art methods.}
\label{tab:efficiency}
\end{table}

\subsection{Human Evaluation}

We further invite radiologists and doctors with expertise in CXR to enrich evaluations. This evaluation is designed to assess three critical aspects of report generation and summarization: correctness, completeness, and coherence. A total of 50 samples are evaluated, with experts rating each aspect on a scale of 0 to 5, where 0 represents the lowest quality and 5 is the highest. 

Figure \ref{fig:human_eval} presents the visualization of the human evaluation scores, comparing our SLaVA-CXR model against LLaVA-Med for both generation (Gen) and summarization (Sum) tasks. The results demonstrate a clear superiority of SLaVA-CXR across all evaluated dimensions. Regarding correctness, SLaVA-CXR outperforms LLaVA-Med, with particularly notable improvements in the summarization task. Regarding completeness, the SLaVA-CXR model shows consistent superiority, especially in the generation task. Regarding coherence, SLaVA-CXR demonstrates markedly better performance, with the most substantial difference observed in the summarization task.

Importantly, clinical experts provide valuable feedback indicating that SLaVA-CXR generally produces reports with better formatting and writing style that more closely align with professional radiology report standards. This is a crucial improvement over LLaVA-Med, which often fell short in this aspect. Furthermore, our model exhibits enhanced correctness and coherence in presenting anatomical descriptions and identifying abnormalities, which are critical for clinical interpretation.

\begin{figure*}[t]
\scriptsize
\centering
\begin{tabular}{p{1.4cm}|p{4.3cm}|p{4.3cm}|p{4.3cm}}
\toprule
 \bf Methods & \textbf{Example 1}      & \textbf{Example 2}    & \textbf{Example 3}    \\ \midrule
\multirow{7}{*}[-3pt]{\textbf{Ground Truth}}  & Normal cardiomediastinal contours. Right lower lung patchy opacities. Small right pneumothorax. Small right pleural effusion.     & No acute cardiopulmonary abnormality.          & There are diffuse predominantly perihilar airspace opacities with slightly nodular appearance, which are new from prior studies. Superimposed hilar adenopathy is difficult to exclude. No pleural effusion or pneumothorax is seen. The cardiomediastinal contours are within normal limits. No acute osseous abnormality is detected.    \\ \midrule

\multirow{8}{*}[-3pt]{\textbf{LLaVA-Med}}   & \textcolor{red}{The image is a chest X-ray taken on day   1 in the emergency department.}  It shows a large \textcolor{blue}{right-sided pleural effusion}, which is an abnormal accumulation of fluid in the pleural space surrounding   the lungs.  \textcolor{red}{Additionally, there is a small left-sized pleural fluid.} \textcolor{red}{The image   does not show any signs of pneumonia or pneumothorax.} & \textcolor{red}{Chyle leak refers to the leakage of   chyle, a milky fluid containing lymph and fat particles, from the lymphonodus (lymphatic vessel) into the   pleural cavity or other surrounding tissues. This can occur as a complication   of certain surgeries or medical procedures.}       & \textcolor{red}{The image is a chest X-ray taken on   postoperative day 1, which shows hypo-inflated lungs without any evidence of   pleural fluid or pneumaothoraces.} \textcolor{blue}{The heart and mediastinal   contours appear normal, and the cardiac dimensions are within the normal   range.} \textcolor{red}{The right sub-clavicular stent, which was previously present, remains   unchanged.}          \\ \midrule

\multirow{4}{*}[-3pt]{\textbf{SLaVA-CXR}}  & \textcolor{blue}{The chest x-ray demonstrates a   right-sided pneumothorax with complete collapse of the right lung and a small   right pleural effusion. The left lung is clear. The heart size is within   normal limits.}    & \textcolor{blue}{The lungs are unremarkable.}   & \textcolor{blue}{The chest x-ray demonstrates rounded   opacity in the upper zone, suggestive of hilar adenopathy. The heart size is   within normal limits. The lungs are clear with no evidence of pleural   effusion.}             \\ \bottomrule
\end{tabular}
\caption{Qualitative analysis of model outputs. Blue-colored text denotes alignment between the ground truth text and the generated text. Red-colored text denotes unfavorable results.}
\label{tab:qual_analysis}
\end{figure*}

\section{Analysis}
We further provide several analyses to better understand our approach.

\subsection{Quantitative Analysis}
Table \ref{tab:ablation_study} presents a  ablation study of the SLaVA-CXR model, which incorporates a Re$^3$Training pipeline consisting of three stages: Recognition, Reasoning, and Reporting. Notably, we observe a clear and consistent progression in performance from Setting (a), which only includes the Recognition stage, through Setting (b), which adds the Reasoning stage, to our full SLaVA-CXR model incorporating all three stages. For instance, in the MIMIC-CXR Generation task, our full model achieves substantial improvements in ROUGE-L and CIDEr. Similarly, for the IU-Xray Summarization task, we see significant gains in BERTScore and CIDEr. This progression is evident across most metrics for both generation and summarization tasks on both datasets, validating the cumulative benefits of each proposed training stage. These results underscore the effectiveness of our Re$^3$Training approach and the value of the RADEX data synthesis method in enhancing the model's capabilities for medical image analysis and report generation tasks.

\subsection{Inference Efficiency}

Table \ref{tab:efficiency} presents a comprehensive analysis of inference efficiency across different models and tasks. The results demonstrate that SLaVA-CXR outperforms both LLaVA-Med and LLaVA v1.5 in terms of computational efficiency across all tasks.

For the report generation task, SLaVA-CXR exhibits remarkable speed improvements. On the IU-Xray dataset, it achieves an inference time of 4.45 seconds per instance, which is approximately 4.4 times faster than LLaVAv1.5 and 15.1 times faster than LLaVA-Med. The efficiency gap is even more pronounced for the MIMIC dataset, where SLaVA-CXR processes each instance in just 3.32 seconds, outpacing LLaVA-Med by a factor of 5.3.

The summarization task showcases even more impressive efficiency gains. Our SLaVA-CXR processes IU-Xray summarization in 1.10 seconds and MIMIC summarization in 1.26 seconds per instance. These times are faster than both LLaVA variants, with SLaVA-CXR being up to 10.5 times faster than LLaVA-Med.

On average, SLaVA-CXR achieves an inference time of 2.53 seconds across all tasks, which is 2.29 times faster than LLaVAv1.5 and an impressive 6.13 times faster than LLaVA-Med. This substantial improvement in computational efficiency is particularly noteworthy given the complex nature of medical image analysis and report generation tasks.

These results underscore SLaVA-CXR's superior design in balancing model performance with computational efficiency.
The reduction in inference time not only enhances the model's practical applicability in clinical settings where rapid report generation is crucial but also demonstrates the potential for more resource-efficient deployment of AI systems in healthcare environments.

\subsection{Qualitative Analysis}

Table \ref{tab:qual_analysis} presents a qualitative comparison of CXR interpretations generated by LLaVA-Med and SLaVA-CXR across three exemplar cases, revealing significant performance disparities.

In Example 1, SLaVA-CXR accurately identifies a right-sided pneumothorax with complete lung collapse and a small right pleural effusion, closely aligning with the ground truth. LLaVA-Med, however, misinterprets these findings, incorrectly identifying a left-sided pleural effusion and missing the pneumothorax.
Example 2 demonstrates SLaVA-CXR's ability to provide concise, accurate assessments for normal findings, mirroring the ground truth. In contrast, LLaVA-Med generates irrelevant and potentially misleading information, showing a tendency towards hallucination when faced with normal findings.
Example 3 further highlights SLaVA-CXR's superior performance, correctly identifying subtle hilar adenopathy and accurately noting the absence of pleural effusion, while LLaVA-Med focuses on less relevant details and misses key findings.

Throughout, SLaVA-CXR consistently demonstrates more accurate anatomical localization, correctly differentiating between right and left-sided findings. This spatial accuracy is crucial in radiological reporting. Moreover, SLaVA-CXR's outputs generally exhibit a more structured and professional reporting style, closely resembling the language and format used in clinical radiology reports.
These qualitative results complement our quantitative findings, providing compelling evidence of SLaVA-CXR's advanced capabilities in CXR interpretation and report generation. The analysis underscores SLaVA-CXR's enhanced proficiency in producing precise, clinically pertinent CXR interpretations compared to LLaVA-Med.

\section{Conclusion}
\label{sec:concludion}


In this work, we propose SLaVA-CXR, a small-scale language and vision assistant for radiology report automation. Our proposed approach addresses key challenges in developing medical domain LLMs through the innovative Re$^3$Training method and RADEX data synthesis technique. Built on a 2.7B parameter backbone, SLaVA-CXR outperforms larger models while achieving six times faster inference efficiency. This research demonstrates significant improvements in chest X-ray report automation, offering an efficient, privacy-compliant solutiofor medical imaging AI. 

\section*{Limitations}

Despite the promising results, there remains significant room for improving SLaVA-CXR's performance. A critical challenge that persists, common among large language models (LLMs), is the issue of hallucination - the generation of plausible but factually incorrect or unsupported information. In the context of medical reporting, such hallucinations could lead to serious clinical misinterpretations. Our model, while advanced, is not immune to this phenomenon. Future research could dive intensively into developing robust strategies to mitigate hallucinations.

Additionally, SLaVA-CXR, along with other models in this study, is trained on single-image inputs. However, clinical practice often involves multiple views (e.g., frontal and lateral) for CXRs to provide a more complete assessment of a patient's condition. This limitation may restrict the model's ability to detect certain pathologies or anatomical variations that are more apparent in alternative views. Future developments should explore multi-view support for both training and inference to enhance the model's diagnostic capabilities and align more closely with clinical workflows.

Lastly, our current model focuses primarily on CXRs. However, radiology encompasses a wide range of imaging modalities and body systems. Expanding the model's capabilities to other areas of radiology and medical imaging would increase its utility in broader clinical contexts.

\section*{Ethical Considerations}
This research adhered to strict ethical guidelines in handling medical data. Our research uses de-identified clinical notes from MIMIC-CXR, and IU-Xray ensuring patient privacy protection. To access MIMIC-CXR data, researchers have completed necessary training course and signed the data use agreement. In compliance with data protection regulations, we exclusively employed locally hosted Large Language Models (LLMs) for data processing, preventing any unauthorized access or transmission of sensitive information.
 
\bibliography{main}
\newpage
\appendix

\section*{Appendix}
\label{sec:appendix}

\section{Experiment Setup}
\label{sec:setup}
\paragraph{Evaluation Data.} The evaluation datasets contain MIMIC-CXR \cite{johnson2019mimic} and IU-Xray \cite{Dina2016IU-Xray} for the task of radiology report generation and summarization. For MIMIC, for a fair comparison, we use the test split from \citet{johnson2019mimic} for evaluation. Frontal images are employed consistently across both datasets, resulting in 1732 samples from MIMIC-CXR and 3301 samples from IU-Xray.

\paragraph{Task Description.} We perform the evaluation on CXR report generation and summarization. Each radiology report comprises a ``Findings'' section, detailing the radiologist's observations from the images, and an ``Impressions'' section, summarizing these observations for diagnostic interpretation. In {CXR report generation}, the objective is to create the "Findings" section from the provided images. For CXR report summarization, the goal is to formulate the ``Impressions'' section using the ``Findings'' and the corresponding images. 

\paragraph{Baseline Models.} Regarding the evluation of our model with other state-of-the-art models, we choose the models with similar architectures. 
Specifically, we choose 4 open-domain models: LLaVA v0 7B (henceforth referred to as \textbf{LLaVAv0}) \citep{liu2023visual} and LLaVA v1.5 7B (henceforth referred to as \textbf{LLaVAv1.5}) \citep{liu2023improved}. It is noteworthy that both LLaVAv0 and LLaVAv1.5 are constrained to a 7B parameter size, aligning with \textbf{LLaVA-Med}, which is the medical domain-finetuned variant of LLaVAv0. As our proposed model, \textbf{SLaVA-CXR}, is built upon the Phi-2 2.7B language model architecture, we also include two Phi2-based models for a fair comparison, which are \textbf{TinyGPT-V} \cite{yuan2023tinygpt}, and \textbf{LLaVA\_phi} \cite{liu2023visual}. A comprehensive description of the model's vision encoder and large language model backbone is provided in Table \ref{tab:desc_model}.
We used eight A6000 GPUs to train and the training hours reported here can be different on other GPUs.



\section{Evaluation Metrics}

\textbf{BLEU \cite{papineni2002bleu}.} BLEU score is an average of n-gram precision, weighted by the so-called brevity penalty that penalizes short high-precision, but low-recall hypotheses.

\textbf{ROUGE \cite{lin2004rouge}.} 
ROUGE is a similarity metric primarily based on recall, which is widely utilized for the evaluation of summarization tasks. Specifically, we opt for ROUGE-L, which focuses on the longest common subsequence (LCS). This approach extends beyond solely considering recall by incorporating an F-score, the harmonic mean of precision and recall, providing a more balanced measure of performance.

\textbf{METEOR \cite{banerjee2005meteor}.} 
In contrast to the aforementioned metrics, METEOR shows a better correction with human judgment. It alters the calculation of precision and recall by implementing a weighted F-score that considers unigram mapping and introduces a penalty function for inaccuracies in word order. Besides, METEOR uses WordNet to expand the set of synonyms, taking the lexical properties of the words into account (e.g., "like" and "likes" should be counted correctly).

\textbf{BERTScore \cite{zhang2019bertscore}.} 
Unlike traditional metrics which rely on exact token matches, BERTScore calculates similarity scores between tokens in candidate and reference sentences using contextual embeddings.

\textbf{CheXbert labeler vector similarity \cite{smit2020chexbert}.} 
This matrix calculates the cosine similarity of embedding from the CheXbert model, which is trained from large-scale chest radiographs.

\textbf{RadGraph \cite{jain2021radgraph}.} 
The RadGraph model parses radiology reports into graphs containing
clinical entities (references to anatomy and observations) and relations between them. The RadGraph F1 metric computes the overlap in entities and relations separately and then reports their average. Entities
are considered to match if the text spans and assigned types are the same, and relations are matched if their endpoint entities and relation types are the same.

\textbf{RadCliQ \cite{yu2023evaluating}.} 
RadCliQ combines the four investigated metrics: BLEU, BERTScore, CheXbert vector similarity, and RadGraph F1. Also, it considers radiologists' feedback in the matrix, which is reported to have the closest alignment with radiologists’ judgment of report quality. The lower value of RadCliQ, unlike all the other ones used in this work, means better quality of the generated response.

\section{Related works}
\label{sec:related_work}

\subsection{Large Language Vision Model (LLVM)}
\label{sec:rel1}
Similar to how ChatGPT quickly evolved into a multimodal vision and language model, open-source large language models (LLMs) have been a driving force in the development of vision-language models. This progress is evident in models like LLaVA \cite{liu2023visual} and MiniGPT4 \cite{zhu2023minigpt}, as well as subsequent versions such as LLaVA-v1.5 \cite{liu2023improved} and MiniGPT-v2 \cite{chen2023minigpt}. These models have effectively showcased that visual instruction tuning significantly enhances multimodal comprehension abilities. Notably, following the success of phi-2, also referred to as a small language model (SLM), both TinyGPT-V\cite{yuan2023tinygpt} and LLaVA-phi \cite{zhu2024llava} mark a paradigm shift towards cost-effective and powerful models, facilitating research in smaller vision-language models.

\subsection{Training Method for LLVM}
\label{sec:rel3}

LLMs usually benefit from the significant advancements from pretraining on vast amounts of data with unsupervised learning. However, it may not be optimized for a specific domain on a specific task. Supervised finetuning in this case bridges this gap by taking advantage of the general language understanding captured during pre-training and adapting it to a target task by guiding with the labeled data. Recent works have shown that following natural language instructions and completing real-world tasks can effectively improve the zero-shot and few-shot generalization abilities of LLMs \cite{taori2023alpaca, chiang2023vicuna}. Furthermore, this has been expanded into LLVM by \citet{liu2023visual} with 2 staged training methods, concept alignment, and visual instruction tuning. Later, \citet{li2023llava} tested this 2 staged training method in the biomedical domain, by training with biomedical multimodal instruction-following data from PMC-15M biomedical figure and caption pairs.

To address these issues, this study contributes open-source, small-scale multimodal large language model which is built on the Phi-2-2.7B model, specifically tailored for radiology report automation. The training method for the model is a three-stage training process in which we add the final training stage with high-quality CXR images and diverse textual data formats to optimize its performance and applicability in radiological contexts. Our work also aims to resolve privacy concerns and resource constraints in using multimodal large language models in healthcare settings, in order to promote LLVM real-world clinical applications. 

\section{Prompts Templates}

\begin{lstlisting}[style=promptstyle, caption={Prompts template for synthetic clinical note generation.}]
[Instructions]: 
You are an expert medical assistant AI capable of modifying clinical documents to user specifications. You make minimal changes to the original document to satisfy user requests. You never add information that is not already directly stated in the original document. Restructure the given text into a radiology report finding. Remove any information not directly observable from the current imaging study. For instance, remove any patient demographic data, past medical history, or comparison to prior images or studies. 

[Input]: 
<Case Description> 
Linear radiolucencies are noted along left mediastinal borders.
No evidence of pneumothorax or pleural effusion.
Lung fields appear clear.
No concerning bony abnormality identified.
<Case Presentation>
Right sided pleuritic chest pain, shortness of breath, diminished breath sounds at right lower zone.
<Case Discussion>
This is a classical case of pneumomediastinum and its presentation.
Asthma is the most common cause.

[Output]: 
Findings: Chest X-ray demonstrates linear radiolucencies along the left mediastinal borders, suggestive of pneumomediastinum. The lung fields appear clear without evidence of pneumothorax or pleural effusion. No concerning bony abnormalities are identified. The cardiac silhouette is within normal limits.
Impression: Pneumomediastinum present. No other acute cardiopulmonary abnormalities are identified. 

\end{lstlisting}

\begin{table}[]
\footnotesize
\begin{center}

\begin{tabular}{llll}
\hline
            & \textbf{GPU}  & \textbf{\#GPU(s)}    & \textbf{Total TFLOPS} \\ \hline
\textbf{Training} & A6000 & 8 & 289.0 \\
\textbf{Evaluation}  & A5000 & 1 & 27.8\\ \hline
\end{tabular}
\caption{GPU resource for training and evaluation.}
\label{tab:gpu_resource}
\end{center}
\end{table}

\begin{table}[]
\footnotesize
\begin{center}

\begin{tabular}{ll}
\hline
      \textbf{Arguments}      &  \textbf{Values}\\ \hline
max token length & 2048 \\
learning rate scheduler & cosine annealing \\
warmup ratio & 0.03 \\
weight decay & 0 \\
gradient checkpointing & true \\ \hline
\end{tabular}
\caption{Training Arguments. Learning Rate, Epochs, and Train Batch Size are described in the Method section.}
\label{tab:training_arguments}
\end{center}
\end{table}

\end{document}